\documentclass[11pt,a4paper]{article}

\usepackage{arxiv}
\usepackage{hyperref} 
\usepackage[utf8]{inputenc} 
\usepackage{graphicx}
\usepackage{soul}
\usepackage{graphicx}  
\usepackage{epstopdf}
\usepackage[T1]{fontenc}    
\usepackage{url}            
\usepackage{booktabs}       
\usepackage{amsfonts}       
\usepackage{nicefrac}       
\usepackage{microtype}      
\usepackage{lipsum}
\usepackage{supertabular}
\usepackage{multirow}
\usepackage{fontenc}
\title{SAT Based Analogy Evaluation Framework for Persian Word Embeddings}
   
\author{
Seyyed Ehsan Mahmoudi \\ 
Shahid Beheshti University \\
se\_mahmoudi@sbu.ac.ir\\
\and
Mehrnoush Shamsfard \\ 
Shahid Beheshti University \\
m-shams@sbu.ac.ir\\ 
}

\begin{document}

\maketitle

\begin{abstract}
          In recent years there has been a special interest in word embeddings as a new approach to convert words to vectors. It has been a focal point to understand how much of the semantics of the the words has been transferred into embedding vectors. This is important as the embedding is going to be used as the basis for downstream  NLP applications and it will be costly to evaluate the application end-to-end in order to identify quality of the used embedding model. Generally the word embeddings are evaluated through a number of tests, including analogy test.  In this paper we  propose a  test framework for Persian embedding models. Persian is a low resource language and there is no rich semantic benchmark to evaluate word embedding models for this language. In this paper we introduce an evaluation framework including a hand crafted Persian SAT based analogy dataset, a colliquial test set (specific to Persian) and a benchmark to study the impact of various parameters on the semantic evaluation task. 
\end{abstract}
\section{Introduction}
In recent years Word Embedding has been used extensively  in many NLP applications. Specially with vast use of deep learning methods, having an embeddings for words, has been a de-facto standard. This highlights the importance of  embedding methods and in the same time urges for a proper framework for evaluating the embedding models. One wholistic approach is to evaluate the embedding models in an end-to-end eco system as extrinsic evaluation method, which will use different embedding models in the specific application and chooses the best one. But this approach is going to be costly in terms of the required computation power. In addition to all hyper-parameters that need to be tuned in later stages it will be challenging to pin point the end-to-end performance for embedding model. 

Alternatively we can have some intrinsic test frameworks to evaluate the quality of embedding model. Traditionally Analogy, Similarity and Categorization tests has been incorporated to perform such intrinsic model evaluations. Analogy tests try to compare the relation of two pairs of words such as \textit{King($v_1$) to Queen($v_2$) is like Man($v_3$) to ? }. This type of test is solved by finding the nearest neighbor of $v_3 + v_2 - v_1$. Word Similarity tests, on the other hand, try to assign a similarity score to pair of words, which normally is the cosine similarity to the words and compare it to human assigned score. Another category of intrinsic tests are Word Categorization tests which try to run a clustering on the words and evaluate the quality of the formed clusters. The test datasets that have been used so far, are mainly inspired by the initial analogy test set that has been used in original Word2Vec and SGNS papers \cite{Mikolov2013Distributed}. This test set has been challenged that it does not cover many semantic aspects of the language. Plus the fact that almost 50 percent of the dataset is dedicated to  some narrow information such as capital-country and country-currency tests. Although solving such analogies shows the power of the embedding model, but in the same time it does not reflect the quality of the embedding model on more semantic oriented fashion. In this sense, there are arguments that the Google analogy dataset is not as challenging as the SAT (Scholastic Aptitude Test) dataset  \cite{church_2017}. Inspection shows that the SAT analogies are all semantic (not syntactic) and involve relatively complex relations. It has been shown that although the answer of analogy is in the top answers but if we inspect other words in top n  nearest neighbor, they are not relevant choices. For example in the \textit{King to Queen is Like Man to ?}, if we inspect top 10 nearest neighbors we expect the majority to be female nouns but this not true  \cite{church_2017}. SAT questions are solved pretty much like the analogy test but reflect much deeper semantic relations. SAT multiple-choice questions come as a stem which is pair of words and 5 other pairs that we should select the pair that has the same semantic relation as the stem. Example of such tests are shown in Table \ref{tab:sat_example}. 
\begin{table} 
  \caption{SAT example question} 
  \centering
  \begin{tabular}{|l r |}
    \hline
    \textbf{Stem} & \textit{mason:stone} \\
    \hline
    a & \textit{teacher:chalk} \\
    b & \textit{carpenter:wood} \\
    c & \textit{soldier:gun} \\
    d & \textit{photograph:camera}\\ 
    e & \textit{book:word} \\
    \textbf{Solution} & \textbf{\textit{carpenter:wood}}\\
    \hline
  \end{tabular}
  \label{tab:sat_example}
\end{table}

The above problem also exists in Persian language model evaluations  where try to simulate the same analogy and word similarity tests on Persian words  \cite{zahedi2018persian, Hadifar2018The}. The analogy test set that is used, has 16 categories of relations where 11 of them are syntactic or morphological relations (See Table \ref{tab:classic_analogy}) \cite{zahedi2018persian}. It does not explore the other aspects and complexity of semantic relations that are crucial in NLP tasks.  In addition to that, Persian language has some special characteristics that are not relevant in other languages such as Colloquialism or colloquial language which has important footprint  in Persian language. In this paper we are going to propose a dataset to evaluate colloquial aspect of the models. We are going to explore the impact of training corpus on performance of the trained models to show how the various aspects of the Persian language will be captured based on the nature of the corpus. 

\section{Analogy Task}
Analogy holds between two word pairs: $a \rightarrow a^* \equiv b\rightarrow b^*$ ($a$ is to $a^*$ as $b$ is to $b^*$) For example, Tokyo is to Japan as Paris is to France. With the pair-based methods, given $a \rightarrow a^* \equiv b\rightarrow ?$, the task is to find $b^*$ from rest of the corpus. An alternative method is to use the set-based methods. With set-based methods, the task is to find $b^*$ given a set of other pairs (excluding $b\rightarrow b^*$) that hold the same relation as $b\rightarrow b^*$. In NLP analogies \cite{Zweig2013Linguistic} are interpreted broadly as basically any "similarities between pairs of words" , not just semantic. See  \cite{church_2017} analysis of Word2Vec, which argues that the Google analogy dataset is not as challenging as the SAT dataset. Inspection shows that the SAT analogies are all semantic (not syntactic) and involve relatively complex relations. In \cite{Jurgens2012SemEval-2012} we can see a taxonomy of the relations that are used in GRE exam and are quite extensive in terms of covering various relationship types. Table \ref{tab:taxonomy} shows a long list of various semantic relations that can be used in analogy task. When the analogy task in Google dataset has accuracy between 80-97 percent in various categories. For SAT analogies this value drops to less than 10 percent. Which shows a huge gap on complexity and completeness of SAT question set. 
 
It has been shown that solving analogies in SAT is far harder for embedding models \cite{Shnayder2004Combining} \cite{Turney2008The}. Same paper also explores the other top N answers of the analogy questions to show that other top 10 answers are not as impressive of the correct answer that we are looking for. This is important as in the real applications we don't know the correct answer and if we look at top N and get a good result, what is the guarantee for the first result to be the best one. In English language we know that solving  SAT questions is considered to be hard problem  that human weighted voting has around 80 percent of accuracy \cite{Lofi2013Just}.

In this paper we have constructed a SAT test benchmark for Persian language. The categories and taxonomy of the relations are kept but the test data set is rebuilt from scratch to reflect true and genuine relations in the words. As the majority of the semantic relations are quite deep, simple translation will definitely fail to produce acceptable outcome. 

\begin{table*}

    \caption{Semantic relations taxonomy (Note that Persian examples are Left to Right)}
     \centering
     \small
     \begin{tabular}{lll}
       \toprule
       Class     &  Sub-Class &  English Example   \\
       \midrule
       Class Inclusion & Taxonomic & emotion:rage  \\
       Class Inclusion & Functional & weapon:knife \\
       Class Inclusion & Singular Collective & clothing:shirt \\
       Class Inclusion & Class Individual & mountain:Everest \\
       Part-Whole & Object:Component & car:engine \\
       Part-Whole & Collection:Member & forest:tree \\
       Part-Whole & Mass:Portion & water:drop \\
       Part-Whole & Event:Feature & wedding:bride \\
       Part-Whole & Activity:Stage & shopping:buying \\
       Part-Whole & Item:Topological Part & mountain:foot \\
       Part-Whole & Object:Stuff & salt:sodium \\
       Part-Whole & Creature:Possession & robin:nest \\
       Part-Whole & Item:Distinctive Nonpart & horse:wings \\
       Part-Whole & Item:Ex-part/Ex-possession & apostate:belief  \\
       Similar & Synonymity & buy:purchase \\
       Similar & Dimensional Similarity  & stream:river \\
       Similar & Dimensional Excessive & concerned:obsessed \\
       Similar & Dimensional Naughty – & copy:plagiarize \\
       Similar & Conversion & apprentice:master \\
       Similar & Attribute Similarity & painting:movie  \\
       Similar & Coordinates & son:daughter \\
       Contrast & Contradictory & masculinity:femininity \\
       Contrast & Contrary & thin:fat \\
       Contrast & Reverse & buy:sell \\
       Contrast & Directional & front:back  \\
       Contrast & Defective & fallacy:logic \\
       Attribute & Item:Attribute (noun:adjective) & soldier:wounded \\
       Attribute & Object Attribute:Condition (adjective:adjective) & brittle:broken \\
       Attribute & Agent Attribute:State (adjective:noun) & contentious:quarrels \\
       Attribute & Object:Typical Action (noun:verb) & glass:break \\
       Attribute & Agent/Object Attribute:Typical Action (adjective:verb) & mutable:change \\
       Non-Attribute & Item:Non-Attribute (noun:adjective) & harmony:discordant \\
       Non-Attribute & Object Attribute:Noncondition (adjective:adjective) & exemplary:criticized \\
       Non-Attribute & Object:Nonstate (noun:noun) & famine:plenitude \\
       Non-Attribute & Attribute:Nonstate (adjective:noun) & immortal:death \\
       Non-Attribute & Object:Atypical Action (noun:verb) & recluse:socialize \\
       Case Relations & Agent:Object & tailor:suit \\
       Case Relations & Agent:Recipient & doctor:patient \\
       Case Relations & Agent:Object - Raw Material &baker:flour \\
       Case Relations & Action:Object & tie:knot \\
       Case Relations & Action:Recipient & teach:student \\
       Case Relations & Object:Recipient & speech:audience \\
       Case Relations & Object:Instrument & violin:bow \\
       Cause-Purpose & Cause:Effect & joke:laughter \\
       Cause-Purpose & Cause:Compensatory Action & hunger:eat \\
       Cause-Purpose & Instrument:Intended Action & gun:shoot \\
       Cause-Purpose & Cause-Purpose Enabling Agent:Object  & car:gas \\
       Cause-Purpose & Action/Activity:Goal & Education:Learning \\ 
       Space-Time & Item:Location & arsenal:weapon \\
       Space-Time & Location:Process/Product & bakery:bread \\
       Space-Time & Location:Action/Activity & school:learn \\
       Space-Time & Time:Action/Activity & summer:harvest \\
       Space-Time & Attachment & belt:waist \\
       Reference & Sign:Significant & siren:danger \\
       Reference & Expression & smile:friendliness \\
       Reference & Plan & agenda:meeting \\
       Reference & Knowledge & psychology:mind \\
       Reference & Concealment & code:meaning \\
       
       \bottomrule
     \end{tabular}
     \label{tab:taxonomy}
   \end{table*}

\subsection{Colloquial Relations}
 Another aspect of building a good benchmark for Persian language is to consider the specific language aspects and reflect them in our test data. One of the specific aspects that we have to pay special attention is the difference in Colloquial and written form of Persian language.  The difference between colloquial and written Persian is much deeper than the difference between colloquial and written English \cite{ghomeshi201812,moradi2018contrastive}. In recent years by wide-spread of social media in Persian speaking society there has been a significant shift in  Colloquial and written form of Persian \cite{article.388818}. Almost no NLP application can work without supporting the Colloquial form in online applications. This leads us to the fact that for the evaluation framework we need to consider this aspect of the language which is missing from main stream NLP datasets on English language. 

\section{Previous Work}
Unfortunately there has not been much done on exploration of embedding models for Persian.In \cite{zahedi2018persian, Hadifar2018The} the performance models are assessed in Analogy, Word Similarity and Categorization tasks. As in this paper we are focusing on analogy based evaluation methods, we will have a closer look at analogy datasets that has been used.  The test sets that has been used for analogy, assess the model in categories that are mentioned in \ref{tab:classic_analogy}. As it is clear very little attention has been paid to finer semantic details of the language also the Colloquial aspect is completely missing from the dataset. 

\begin{table}
  \caption{Classic Relations in Analogy Dataset}
  \centering
  \begin{tabular}{|c c |}
    \hline
    Relationship Type & Number \\
    \hline
    Family Relationship & 342 \\
    Currency & 1260 \\
    Country-Capital & 5402 \\
    Province-Capital & 7832\\ 
    Adjective-Adverb & 1332 \\
    Noun-Adverb & 1056\\
    Antonym & 1260 \\
    Comparative & 1260\\
    Superlative & 1260 \\
    Nationality & 1406\\
    Singular-Plural & 2550 \\
    1st Person & 1260\\
    3rd Person & 1332 \\
    Infinitive-Past & 1260\\
    Infinitive-Present & 1260\\ 

    \hline
  \end{tabular}
  \label{tab:classic_analogy}
\end{table}
\section{The proposed framework}
In this paper we are trying to find a more semantic oriented framework for evaluating the embedding models in Persian language. We first describe the process of the test data creation, which is basically a two step process. We first build a categorical related words dictionary and for the second step  we extract a pool of questions based on the categorical relational dictionary. Once the dataset is ready we will benchmark the result on a wide range of models with various hyper parameters to assess and analyse the framework. 

\subsection{Categorical Relational Dictionary}

The first step in our SAT framework is to create the questions. As for English language  there is a long history of exams such as GRE(Graduate Record Examinations), building such datasets is much easier \cite{TurneySAT}. In Persian on the other hand we chose not to translate the English datasets. As the relations have deep semantic aspects, translation becomes misleading.This is due to the fact that deep semantic relations normally rely on specific sense of the words and for majority of the words, building a one to one relation between the words in two languages is not possible or practical. Furthermore disambiguation of word senses need them to be in context, while analogy tests use separate words and so using WSD methods before translation is not possible. On the other hand due to some lexical, conceptual or cultural gaps, we may not find suitable Persian correspondence for each English pair in the dataset.
For this reason, we build a new hand crafted categorical word dictionary from scratch for Persian Language. We based our work on  relational categories proposed for SemEval2017 \cite{Jurgens2012SemEval-2012}. The relational categories are quite cross lingual and are properly relevant to Persian language.  We create word pairs as dictionary entries belonging to all of the relational categories. We created word pairs  by human experts and we required at least 5 pairs per each category.  We build 390 word pairs that are organized in 67 relational categories mentioned in Table \ref{tab:taxonomy}. Those are the same catagories that are specified in \cite{semeval_cats}. Because in  our  tokenization methods we are tokenizing based on spaces, our models are incapable of working with compound verbs. For this limitation, for compound verbs we replaced them with their infinitive which is just one token. This keeps the semantic context of the word and in the same time overcomes the tokenization limitation. On the same note for some of the words we did not stick to POS of words that are in the table and  focused on the semantic relation that is represented by the categories. Again this was due to semantic aspects of the Persian languages that we take into consideration.

\subsubsection{Building the Test Pool}
The next step is to build the test pool based on the categorical word pairs. We use a randomized algorithm to create a comprehensive question pool. Each 5-choice question has a stem  which is a word pair and 5 answer options which are word pairs as well. Each question has just one correct answer and the word pair in correct answer option is in the same relational category of the stem word pair and other pairs are selected from other categories. Using randomized algorithm we generate questions.  As mentioned above the actual number of word pairs in our dataset is 390 but the combinations of word pairs against other categories will make a much larger space for questions to be generated. 
 
 We  prepared a question set  for automatic evaluation of the models containing 5000 questions. From this dataset we use 1000 randomly drawn subset for human cross validation. The human cross validation dataset is to assess the human ability to solve such test.  The reason to use a subset of question was to meet the attention span of our audience. We reduced the test size for each person to 20 questions.  For each participant in our experiment we sequentially provided the questions to them and collected the answers. 
  
For each question we have a pair of words as stem, $a \rightarrow a^*$ and a number (5) of options $o_i \rightarrow o_{i}^*$. We simply calculate  vectors  $a^* - a$ and options vectors $o_{i}^* - o_i$ and calculate the cosine similarity of options vectors with stem vector. The option with smallest distance is considered to be the answer of the question.  In this task the base-line performance is random choice that has 20 percent chance to be correct. So the random baseline is 20 percent. The algorithm is quite straight forward and no hyperparameter exists. 

 \subsection{Colloquial Analogy}
 Also to explore the behavior of embedding models in colloquial aspects of language, specific dataset has been prepared that only contains such pairs of words.As there is no formal rules to convert words to their colloquial counterpart and also the meaning is the same on two words this relation is treated specially.  The analogy data that has been prepared has 2332  analogy questions and it is as the same form as traditional analogy datasets. We intentionally created a separate dataset for this part as it does not fit into the SAT questions category. 

\section{Experimental Setup}

 \subsection{Training Corpora}
 We want to explore impact of the base corpus on performance of the model on two tasks (SAT Analogy and Colloquial Analogy) that are introduced above. We use corpora from Persian Wikipedia, Persian Blog Corpus(hmBlog) \cite{hmBlog} and  Persian Twitter Corpus. Each of the mentioned sources is representing various aspects of Persian Language. In Table \ref{tab:corpora} you can find more details on the corpora that is used for training. All of the above corpora are normalized and cleaned up. The characters not in Persian alphabet are removed (except the numeric characters). The multi-form characters that are similar in written form but different unicode characters, has been transformed to their standard Persian Unicode character. This normally happens once the original texts are typed in Arabic keyboard layouts. For this reason the training corpora  are normalized.    

 \begin{table}
  \caption{Training Corpora Details}
  \centering
  \begin{tabular}{|l | l | l |}
    \hline
    Source &Sentences & Token Count \\
    \hline
    Wikipedia & 5M & 55M\\ 
    Persian Blogs & 400+M & 5.4 Billion \\
    Persian Twitter & 70M & 930M \\ 
    \hline
  \end{tabular}
  \label{tab:corpora}

\end{table}

\subsection{Model Training}
There are various methods to train the embedding model. we will be examining FastText (CBOW and Skipgram with Negative Sampling SGNS) methods \cite{bojanowski2016enriching}. FastText used as a representative of many methods that are used in the literature. For all of them   we are exploring dimensions (50, 100 , 300 , 400) and window size (3, 5, 10). Considering 3 corpora that will be used for training, it ended up in 72 trained models. On each model  3 tasks are running, normal analogy test\cite{zahedi2018persian}, SAT-like analogy test and colloquial analogy test. 
 
\section{Experiment Results}
Table \ref{tab:analogy_result} outlines the results of all three  tasks on the trained models. There is a significant difference in the results of the various models. This shows the significant impact of both corpus and the hyper parameters that we have to choose on the final outcome. 
\begin{table} 
  \caption{Classic Analogy Result}
  \centering
  \tiny
  \begin{tabular}{|l | l | l | l | l | l | l | l | }
    \hline
    Model & Corpus & D & W & SAT & Analogy &  Colloquial  \\
    \hline
    \multirow{27}{*}{CBOW} & \multirow{9}{*}{Wikipedia} & \multirow{3}{*}{100} & 3&  0.385 &  0.290 &  0.068\\ \cline{4-7}
                          &                            &                      & 5 &  0.374 &  0.277 &  0.063\\ \cline{4-7}
                          &                            &                      & 10&  0.348 &  0.270 &  0.060\\ \cline{3-7}
                          &                            & \multirow{3}{*}{300} & 3 &  0.394 &  0.258 &  0.093\\ \cline{4-7}
                          &                            &                      & 5 &  0.390 &  0.262 &  0.078\\ \cline{4-7}
                          &                            &                      & 10 & 0.374 &  0.251 &  0.076\\ \cline{3-7}
                          &                            & \multirow{3}{*}{400} & 3 &  0.391 &  0.233 &  0.087\\ \cline{4-7}
                          &                            &                      & 5 &  0.387 &  0.231 &  0.083\\ \cline{4-7}
                          &                            &                      & 10 & 0.356 &  0.232 &  0.073\\ \cline{2-7}
                          & \multirow{9}{*}{Blogs}     & \multirow{3}{*}{100} & 3 &  0.421 &  0.301 &  0.431\\ \cline{4-7}
                          &                            &                      & 5 &  0.407 &  0.296 &  0.386\\ \cline{4-7}
                          &                            &                      & 10 & 0.396 &  0.309 &  0.332\\ \cline{3-7}
                          &                            & \multirow{3}{*}{300} & 3 &  0.439 &  0.356 &  0.450\\ \cline{4-7}
                          &                            &                      & 5 &  0.424 &  0.350 &  0.401\\ \cline{4-7}
                          &                            &                      & 10 & 0.413 &  0.362 &  0.342\\ \cline{3-7}
                          &                            & \multirow{3}{*}{400} & 3 &  0.436 &  0.350 &  0.450\\ \cline{4-7}
                          &                            &                      & 5 &  0.428 &  0.355 &  0.399\\ \cline{4-7}
                          &                            &                      & 10 & 0.417 &  0.376 &  0.345\\ \cline{2-7}
                          & \multirow{9}{*}{Twitter}   & \multirow{3}{*}{100} & 3  & 0.360 &  0.183 &  0.396\\ \cline{4-7}
                          &                            &                      & 5  & 0.344 &	0.184	& 0.364\\ \cline{4-7}
                          &                            &                      & 10 & 0.342 &	0.183	& 0.343 \\ \cline{3-7}
                          &                            & \multirow{3}{*}{300} & 3 & 0.382 &  0.180 &  0.375  \\ \cline{4-7}
                          &                            &                      & 5 & 0.373 &  0.186 &  0.363 \\ \cline{4-7}
                          &                            &                      & 10 & 0.366 &  0.178 &  0.331  \\ \cline{3-7}
                          &                            & \multirow{3}{*}{400} & 3 & 0.387 &  0.176 &  0.351  \\ \cline{4-7}
                          &                            &                      & 5 & 0.379 &  0.172 &  0.336  \\ \cline{4-7}
                          &                            &                      & 10 & 0.366 &  0.171 &  0.315 \\ \cline{1-7}
\multirow{27}{*}{SGNS} & \multirow{9}{*}{Wikipedia} & \multirow{3}{*}{100} & 3 & 0.379 &  0.397 &  0.083\\ \cline{4-7}
                          &                            &                      & 5 & 0.373 &  0.395 &  0.064\\ \cline{4-7}
                          &                            &                      & 10 & 0.363 &  0.377 &  0.040\\ \cline{3-7}
                          &                            & \multirow{3}{*}{300} & 3 & 0.395 &  0.370 &  0.099\\ \cline{4-7}
                          &                            &                      & 5 & 0.388 &  0.394 &  0.103\\ \cline{4-7}
                          &                            &                      & 10 & 0.368 &  0.418 &  0.064\\ \cline{3-7}
                          &                            & \multirow{3}{*}{400} & 3 & 0.397 &  0.315 &  0.080\\ \cline{4-7}
                          &                            &                      & 5 & 0.388 &  0.348 &  0.077\\ \cline{4-7}
                          &                            &                      & 10 & 0.384 &  0.405 &  0.075\\ \cline{2-7}
                          & \multirow{9}{*}{Blogs}     & \multirow{3}{*}{100} & 3 &  0.434 &  0.406 &  0.548\\ \cline{4-7}
                          &                            &                      & 5 & 0.428 &  0.412 &  0.506\\ \cline{4-7}
                          &                            &                      & 10 & 0.410 &  0.411 &  0.451\\ \cline{3-7}
                          &                            & \multirow{3}{*}{300} & 3 & \textbf{ 0.473} &  0.457 &  \textbf{0.602}\\ \cline{4-7}
                          &                            &                      & 5 &  0.453 &  0.463 &  0.582\\ \cline{4-7}
                          &                            &                      & 10 & 0.434 &  0.479 &  0.549\\ \cline{3-7}
                          &                            & \multirow{3}{*}{400} & 3 & 0.472 &	0.466	 & 0.578\\ \cline{4-7}
                          &                            &                      & 5 & 0.441 &  0.467 &  0.567\\ \cline{4-7}
                          &                            &                      & 10 & 0.437 &  \textbf{0.481} &  0.533\\ \cline{2-7}
                          & \multirow{9}{*}{Twitter}   & \multirow{3}{*}{100} & 3 & 0.361 &  0.250 &  0.515\\ \cline{4-7}
                          &                            &                      & 5 & 0.350 &  0.255 &  0.474\\ \cline{4-7}
                          &                            &                      & 10 & 0.342 &  0.251 &  0.458\\ \cline{3-7}
                          &                            & \multirow{3}{*}{300} & 3 &  0.404 &  0.218 &  0.414\\ \cline{4-7}
                          &                            &                      & 5 &  0.385 &  0.212 &  0.437\\ \cline{4-7}
                          &                            &                      & 10 & 0.373 &  0.231 &  0.444\\ \cline{3-7}
                          &                            & \multirow{3}{*}{400} & 3 & 0.394 &  0.193 &  0.333\\ \cline{4-7}
                          &                            &                      & 5 & 0.394 &  0.190 &  0.341\\ \cline{4-7}
                          &                            &                      & 10 & 0.380 &  0.202 &  0.370\\ \cline{2-7}
    \hline
  \end{tabular}
  \label{tab:analogy_result}
\end{table}

\subsection{Analogy Task}
Here also  the impact of the corpus is clear and simply can be stated that the larges corpus results in the better result. The best result for Analogy (\textbf{0.48}) improves the previous benchmarks by  \textbf{3\%} \cite{zahedi2018persian} which is clearly the impact of the larger corpus used in this article. The result of the less diverse corpora (Wikipedia and Twitter) are significantly lower compared to hmBlog corpus we used. Same as what has been reported before SGNS outperforms the CBOW models \cite{zahedi2018persian, Hadifar2018The}.

 In analogy task the variation is much less compared to other tasks. This signifies that the test itself  is not giving us a proper indication of the quality of the model in subsequent tasks. As the variation is high and values are quite close to each other. 

 \subsection{Colloquial}
For Colloquial analogy task, it is clear that formal text resources for corpus such as Wikipedia are performing really poor. The highest accuracy achieved on Wikipedia corpus was \textbf{0.10} compared to other models that achieved up to \textbf{0.60} for Colloquial analogy task. This shows the importance choosing proper corpus to capture specific aspects of the language. 

\subsection{SAT Test}
In the SAT test  it is clear that the result is heavily dependent on the quality of the corpus and parameters. Interesting point is that majority of the output are having almost the same accuracy but for specific models the outcome is significantly better. This shows the fact that this test can be a good discriminatory measure for various embedding models. Even on a rich corpus such as hmBlog, the diversity of the results is quite notable. The best result belongs to SGNS model trained on hmBlogs corpus with dimension 300 and window size 3. Although for both Analogy and SAT the best corpus is the hmBlog corpus but the best result comes out of a different hyper parameters.  It is interesting that the best result for SAT and Colloquial Analogy are based on the same model. Although the colloquial analogy test has the same format as the classic analogy test, but as it is more inclined to the semantic aspect of the language, the colloquial analogy results show more correlation with SAT compared to classic analogy. Table \ref{tab:corrolation} shows that SAT test has the highest mutual correlation of the results with the other methods.

\begin{table}
  \caption{Tasks correlation Analysis}
  \centering
  \begin{tabular}{|l | c |}
    \hline
    SAT - Analogy & 0.37 \\
    \hline
    Colloquial - Analogy & 0.20 \\ 
    \hline
    SAT - Colloquial & 0.59 \\
    \hline
  \end{tabular}
  \label{tab:corrolation}
\end{table}

\section{Human SAT Analogy Test}
In order to find the difficulty level of the task of analogy for Persian language we developed a test website \footnote{\url{https://sbu-nlp-sat.herokuapp.com/}} to conduct the test by human and evaluate the general difficulty of the task.  For this experiment, as mentioned before, 1000 random questions formed the main question bank. We provided 20 questions to each participant. The data was collected and stored in database.  Once  participants start the test, they are given the next batch of the questions. We did not perform any randomization here to make sure that all questions will be answered at least once. 
 
\begin{figure}
  \centering
\includegraphics{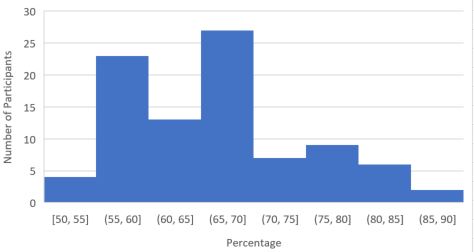}
\caption{Human Voting Result Histogram}
\label{pic:human_histo}  
\end{figure}

In total 94 participants took part in the test. The average result of all participants was 68 percent with median of 70 percent. In Figure \ref{pic:human_histo} you can find the histogram of the result of the human answers. This quite resides in the same range of English SAT test historical result which is 57 percent \cite{article.388818}. This indicates the complexity and ambiguity of the task even for human. In the same time outlines the richness of the embedding models that have achieved quite human comparable outcome.

\section{Conclusion}
The results of various tasks specific for Persian language demonstrate that special attention needs to be made for characteristics of the language being processed. Also we see that the coverage of corpus plays important role on performance of the embedding models in the tasks. As shown the SAT test result had better mutual correlation with the other two test datasets. This highlights that it can be used as good measure of the model quality. Introduction of the SAT based analogy task, helps us assess the depth of the semantic richness of the embedding model before jumping into broader end-to-end application. Looking at the results of human performance versus the embedding model performance signifies the maturity and richness of the embedding models. Also the dataset created in this research can act as a benchmark on Persian language, whilst unfortunately such datasets are rare. For next steps we are working to extend the dataset beyond the current hand crafted list of words to more automated way of extracting deep semantic relations. One of the good candidates that we are working on is the FarsNet \cite{FarsNets}. This  helps us to develop much broader dataset with larger words in it.

\bibliographystyle{apalike} 
\bibliography{references}

\begin{thebibliography}{}

\bibitem[Bojanowski et~al., 2016]{bojanowski2016enriching}
Bojanowski, P., Grave, E., Joulin, A., and Mikolov, T. (2016).
\newblock Enriching word vectors with subword information.
\newblock {\em arXiv preprint arXiv:1607.04606}.

\bibitem[Church, 2017]{church_2017}
Church, K.~W. (2017).
\newblock Word2vec.
\newblock {\em Natural Language Engineering}, 23(1):155–162.

\bibitem[Ghomeshi, 2018]{ghomeshi201812}
Ghomeshi, J. (2018).
\newblock 12 the associative plural and related constructions in persian.
\newblock {\em Trends in Iranian and Persian Linguistics}, 313:233.

\bibitem[Hadifar and Momtazi, 2018]{Hadifar2018The}
Hadifar, A. and Momtazi, S. (2018).
\newblock The impact of corpus domain on word representation: a study on
  persian word embeddings.
\newblock {\em language resources and evaluation}, 52(4):997--1019.

\bibitem[Hamzeh and Chen, 2018]{moradi2018contrastive}
Hamzeh, M. and Chen, J. (2018).
\newblock A contrastive analysis of persian and english vowels and consonants.
\newblock {\em Lege Artis}, 3(2):105--131.

\bibitem[Jurgens et~al., 2012a]{Jurgens2012SemEval-2012}
Jurgens, D., Holyoak, K.~J., Mohammad, S.~M., and Turney, P.~D. (2012a).
\newblock Semeval-2012 task 2: Measuring degrees of relational similarity.
\newblock {\em joint conference on lexical and computational semantics},
  1:356--364.

\bibitem[Jurgens et~al., 2012b]{semeval_cats}
Jurgens, D.~A., Turney, P.~D., Mohammad, S.~M., and Holyoak, K.~J. (2012b).
\newblock Semeval-2012 task 2: Measuring degrees of relational similarity.
\newblock In {\em Proceedings of the First Joint Conference on Lexical and
  Computational Semantics - Volume 1: Proceedings of the Main Conference and
  the Shared Task, and Volume 2: Proceedings of the Sixth International
  Workshop on Semantic Evaluation}, SemEval '12, page 356–364, USA.
  Association for Computational Linguistics.

\bibitem[Lofi, 2013]{Lofi2013Just}
Lofi, C. (2013).
\newblock Just ask a human? - controlling quality in relational similarity and
  analogy processing using the crowd.
\newblock {\em btw workshops}, pages 197--210.

\bibitem[Mikolov et~al., 2013]{Mikolov2013Distributed}
Mikolov, T., Sutskever, I., Corrado, G.~S., Dean, J., and Chen, K. (2013).
\newblock Distributed representations of words and phrases and their
  compositionality.
\newblock {\em arXiv: Computation and Language}.

\bibitem[Motahari and Shamsfard, 2020]{hmBlog}
Motahari, H. and Shamsfard, M. (2020).
\newblock Extracting metaphorical ajdective phrases based on corpus and word
  embedding models.
\newblock In {\em 2nd National Conference on Applied Research in Computational
  Linguistics}.

\bibitem[Shamsfard et~al., 2010]{FarsNets}
Shamsfard, M., Hesabi, A., Fadaei, H., Mansoory, N., Famian, A., Bagherbeigi,
  S., Fekri, E., Monshizadeh, M., and Assi, M. (2010).
\newblock Semi automatic development of {Farsnet}; the persian wordnet.
\newblock In {\em Proceedings of 5th Global WordNet Conference}, volume~29.

\bibitem[Shnayder et~al., 2004]{Shnayder2004Combining}
Shnayder, V., Littman, M.~L., Bigham, J.~P., and Turney, P.~D. (2004).
\newblock Combining independent modules in lexical multiple-choice problems.
\newblock {\em recent advances in natural language processing}, pages 101--110.

\bibitem[Shohani and Hosseini, 2018]{article.388818}
Shohani, A. and Hosseini, S. (2018).
\newblock The impact of cyberspace on contemporary persian language and
  literature.
\newblock {\em Persian Language and Literature}, (237):75--101.

\bibitem[Turney, 2008]{Turney2008The}
Turney, P.~D. (2008).
\newblock The latent relation mapping engine: algorithm and experiments.
\newblock {\em Journal of Artificial Intelligence Research}, 33(1):615--655.

\bibitem[Turney et~al., 2003]{TurneySAT}
Turney, P.~D., Littman, M.~L., Bigham, J., and Shnayder, V. (2003).
\newblock Combining independent modules to solve multiple-choice synonym and
  analogy problems.
\newblock {\em CoRR}, cs.CL/0309035.

\bibitem[Zahedi et~al., 2018]{zahedi2018persian}
Zahedi, M.~S., Bokaei, M.~H., Shoeleh, F., Yadollahi, M.~M., Doostmohammadi,
  E., and Farhoodi, M. (2018).
\newblock Persian word embedding evaluation benchmarks.
\newblock In {\em Electrical Engineering (ICEE), Iranian Conference on}, pages
  1583--1588. IEEE.

\bibitem[Zweig et~al., 2013]{Zweig2013Linguistic}
Zweig, G., tau Yih, W., and Mikolov, T. (2013).
\newblock Linguistic regularities in continuous space word representations.
\newblock {\em north american chapter of the association for computational
  linguistics}, pages 746--751.

\end{thebibliography}

\end{document}